\documentclass[conference]{IEEEtran}

\usepackage{cite}
\usepackage{graphicx}
\usepackage{cite}
\usepackage{picinpar}
\usepackage{amsmath}
\usepackage{stfloats}
\usepackage{url}
\usepackage{flushend}
\usepackage{colortbl}
\usepackage{soul}
\usepackage{multirow}
\usepackage{pifont}
\usepackage{color}
\usepackage{xcolor}
\usepackage[process=auto]{pstool} 
\usepackage{enumerate}
\usepackage{footnote}
\usepackage{graphicx,import}
\usepackage{graphicx}
\usepackage{adjustbox}
\usepackage{float}
\usepackage[caption=false,font=normalsize,labelfont=sf,textfont=sf]{subfig}
\usepackage{alltt}
\usepackage[hidelinks]{hyperref}
\usepackage{enumerate}
\usepackage{siunitx}
\DeclareSIUnit \var { Var }
\usepackage{breakurl}
\usepackage{epstopdf}
\usepackage{pbox}
\usepackage{import}
\usepackage{calc}
\usepackage{algorithm}
\usepackage{algorithmicx}
\usepackage{algpseudocode}
\usepackage{amsthm}
\usepackage{amsmath}
\usepackage{amssymb} 
\usepackage{etoolbox} 
\appto\appendix{\counterwithin{equation}{section}} 
\def\BibTeX{{\rm B\kern-.05em{\sc i\kern-.025em b}\kern-.08em
    T\kern-.1667em\lower.7ex\hbox{E}\kern-.125emX}}
 \usepackage{tabularray}
\usepackage{tabularx,booktabs}
\newcolumntype{C}{>{\centering\arraybackslash}X} 
\usepackage{lipsum}
\def\BibTeX{{\rm B\kern-.05em{\sc i\kern-.025em b}\kern-.08em
    T\kern-.1667em\lower.7ex\hbox{E}\kern-.125emX}}
\usepackage{amsfonts} 
\usepackage{pifont}

\begin{document}

\title{Uncertainty-Aware Federated Learning for Cyber-Resilient Microgrid Energy Management \\
\thanks{This work was partly supported by Innovative Human Resource Development
for Local Intellectualization program through the Institute of IITP grant
funded by the Korean government(MSIT) (IITP-2025-RS-2020-II201612,
50\%) and by Priority Research Centers Program through the NRF funded
by the MEST(2018R1A6A1A03024003, 50\%).}
}

\author{Oluleke Babayomi, Dong-Seong Kim}

\author{\IEEEauthorblockN{Oluleke Babayomi}
\IEEEauthorblockA{\textit{ICT Convergence Research Center} \\
\textit{Kumoh National Institute of Technology}\\
Gumi, South Korea \\
babayomi@ieee.org}
\and
\IEEEauthorblockN{Dong-Seong Kim }
\IEEEauthorblockA{\textit{IT-Convergence Engineering} \\
\textit{Kumoh National Institute of Technology}\\
Gumi, South Korea \\
dskim@kumoh.ac.kr}
}

\maketitle
\begin{abstract}
Maintaining economic efficiency and operational reliability in microgrid energy management systems under cyberattack conditions remains challenging. Most approaches assume non-anomalous measurements, make predictions with unquantified uncertainties, and do not mitigate malicious attacks on renewable forecasts for energy management optimization. This paper presents a comprehensive cyber-resilient framework integrating federated Long Short-Term Memory-based photovoltaic forecasting with a novel two-stage cascade false data injection attack detection and energy management system optimization. The approach combines autoencoder reconstruction error with prediction uncertainty quantification to enable attack-resilient energy storage scheduling while preserving data privacy. Extreme false data attack conditions were studied that caused 58\% forecast degradation and 16.9\% operational cost increases. The proposed integrated framework reduced false positive detections by 70\%, recovered 93.7\% of forecasting performance losses, and achieved 5\% operational cost savings, mitigating 34.7\% of attack-induced economic losses. Results demonstrate that precision-focused cascade detection with multi-signal fusion outperforms single-signal approaches, validating security-performance synergy for decentralized microgrids.
\end{abstract}


\section{Introduction}

Accurate renewable energy forecasting is fundamental to reliable energy management systems (EMS) in modern power grids. However, these forecasting systems face critical vulnerabilities to false data injection attacks (FDIA), which can severely degrade EMS reliability and operational efficiency. Comprehensive mitigation frameworks must integrate robust anomaly detection with system recovery mechanisms to counteract attack-induced degradation.

Prior research has predominantly focused on sophisticated attack detection techniques. Del Fiore et al. developed an IoT platform combining real-time data acquisition with ensemble machine learning (SVR, Random Forest, Ridge Regression) for solar forecasting and automated anomaly notifications \cite{DelFiore2025}. Guarino et al. employed Meta's Prophet forecasting model with residual-based anomaly detection, achieving R$^2$ of 0.979 for large-scale diagnostics \cite{Guarino2025}. In cyber-physical systems, \cite{Wang2019f} proposed interval state forecasting with kernel quantile regression to detect dynamic injection attacks, while \cite{Roy2023} developed model-based intrusion detection for automatic generation control systems accounting for system uncertainties. Feature selection optimization further advanced detection: binary particle swarm-wrapped selection framework (BPSWO) achieved state-of-the-art accuracy on IEEE test systems \cite{Han2023}, and \cite{Zhang2023b} proposed cyber-physical symmetry detection using unknown input observers.
 
 Despite these advances, critical limitations persist. Classical methods—Local Outlier Factor (LOF), DBSCAN, and Isolation Forest—suffer fundamental limitations: LOF exhibits O(n²) computational complexity with high hyperparameter sensitivity; DBSCAN struggles with parameter tuning ($\epsilon$, min\_samples) and varying-density regions; Isolation Forest fails to detect collective temporal anomalies. Crucially, existing approaches largely overlook: (1) uncertainty quantification in forecasting models, leading to confident predictions despite model uncertainty; (2) privacy-preserving collaborative learning in distributed microgrids; and (3) end-to-end economic validation of mitigation effectiveness. Meanwhile, our prior work has demonstrated cyber-resilient forecasting for smart grid infrastructure with anomaly mitigation \cite{Babayomi2025a}, and \cite{Gal2016} has proposed computationally efficient Monte Carlo Dropout uncertainty quantification via post-hoc uncertainty refinement.

Therefore, this paper addresses these research gaps for renewable forecasting through: (1) a novel two-stage cascade anomaly detection combining autoencoder reconstruction error with Monte Carlo Dropout uncertainty quantification to reduce false alarms; (2) a privacy-preserving federated learning architecture enabling decentralized attack mitigation across distributed prosumers; (3) quantified demonstration of both forecasting performance recovery and operational cost mitigation under FDIA conditions.

The paper is organized as follows. Section \ref{Sec:SysArchi} introduces the system architecture, and Section \ref{Sec:ProposeMethodology} describes the proposed method. Sections \ref{Sec:ExpSetup}, \ref{Sec:ResultsNAnalysis}, \ref{Sec:Discussion} and \ref{Sec:Conclusion} cover the experimental setup, results, discussion and conclusion, respectively.

\begin{figure*}[t]
    \centering    \includegraphics[width=0.95\linewidth]{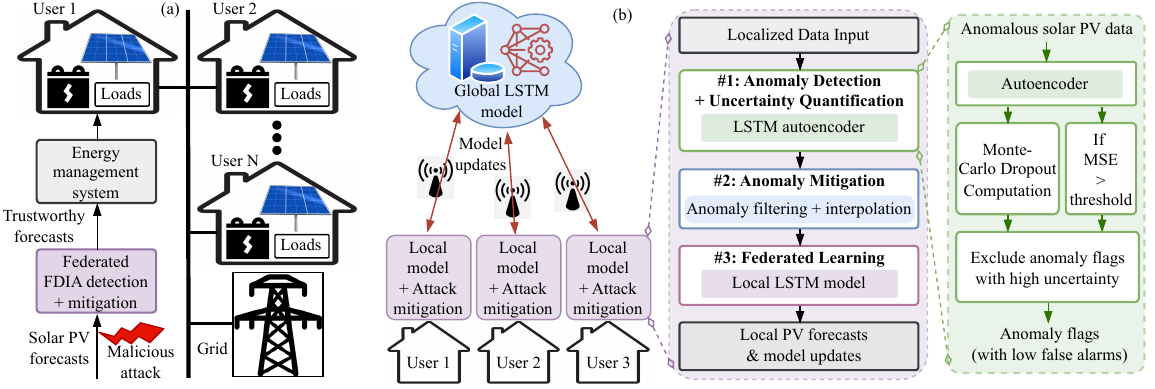}
    \caption{System diagram. (a) Overall system architecture. (b) Proposed federated learning approach to FDIA-resilient PV forecasting and energy management.}
    \label{fig:OverallDiag}
\end{figure*}

\section{System Architecture and Threat Model}
\label{Sec:SysArchi}
Figure \ref{fig:OverallDiag} shows several houses, each representing a user. Each house is equipped with a Solar PV panel (for generation), energy storage, and electrical loads. They are connected to a central electrical grid. Solar PV forecast data is subject to a malicious attack (FDIA), which aims to corrupt the integrity of the forecast. A novel attack detection and data cleaning process is proposed in this study using a federated learning approach enhancing privacy and robustness. The output of the detection and mitigation module will produce trustworthy solar PV forecasts for the EMS optimal management of the microgrid's energy storage scheduling. 

\subsection{Federated Learning Architecture}

The proposed system comprises three geographically distributed users. Each user operates a local PV installation with battery energy storage managed by an EMS determining optimal charging schedules from day-ahead forecasts.

The federated learning architecture enables collaborative model training without centralized data collection. Each user maintains a local LSTM-based forecasting model trained on historical PV generation. At regular intervals, local model parameters transfer to a central aggregator executing the Federated Averaging (FedAvg) algorithm, distributing updated global parameters to all participants.

The LSTM architecture employs sequential layers capturing temporal PV generation patterns. Hyperparameters include: sequence length of 24 time steps (capturing diurnal patterns), 64 LSTM units, learning rate of 0.001, 10 federated rounds, and 20 local training epochs per round per user.

\subsection{Threat Model: False Data Injection Attacks}

The adversary model assumes a sophisticated attacker capable of compromising local sensors or data infrastructure at any participating site. FDIA simulations employ realistic attack characteristics: 1) {Attack Pattern:} Targeted manipulation of PV measurements during specific time windows, introducing both zero-injection attacks (simulating sensor failures) and magnitude manipulation (reducing reported generation by 80-90\%). 2)
{Attack Distribution:} Spatially and temporally distributed attacks across all three users with attack probabilities of 15\%, 12\%, and 10\% for Users 1, 2, and 3 respectively, resulting in 50\% overall data compromise during the evaluation window. {Attack Objective:} Persistent degradation of forecast accuracy leading to suboptimal EMS decisions and increased operational costs rather than causing immediate operational disruptions.

\section{Proposed Cyber-Resilient Framework}
\label{Sec:ProposeMethodology}
\subsection{Two-Stage Cascade Anomaly Detection}
The proposed novel anomaly detection technique is in two-stages: autoencoder reconstruction error detection, and novel uncertainty quantification refinement. 
\subsubsection{Stage 1: Autoencoder-Based Reconstruction Error Detection}

Autoencoders are unsupervised neural networks learning compact representations of normal data patterns through bottleneck architectures. Anomalies are detected via reconstruction error, with the hypothesis that malicious data points exhibit significantly higher reconstruction errors than legitimate measurements. 
Each user deploys a local autoencoder trained exclusively on verified clean historical data collected prior to attack periods. The architecture includes and encoder and decoder; an anomaly anomaly threshold $\tau$ is determined through statistical analysis of reconstruction errors on clean validation data, defined as $\tau = \mu + 2.5\sigma$, where $\mu$ and $\sigma$ represent mean and standard deviation of training reconstruction errors.

For each incoming PV measurement $x_t$, the autoencoder computes mean squared error (MSE) reconstruction error:
\begin{equation}
\text{MSE}_t = \frac{1}{n} \sum_{i=1}^{n} (x_t^{(i)} - \hat{x}_t^{(i)})^2
\end{equation}

Measurements with $\text{MSE}_t > \tau$ are flagged as potential anomalies.

\subsubsection{Stage 2: Uncertainty Quantification Refinement}

Stage 2 refines Stage 1 detections using Monte Carlo Dropout uncertainty estimation. The LSTM forecast model executes 50 forward passes with dropout enabled ($p_{dropout} = 0.3$) to estimate prediction confidence:

\begin{equation}
\hat{y}_t^{MC} = \frac{1}{T} \sum_{k=1}^{T} f_k(x_t)
\end{equation}

where $T=50$ and $f_k$ denotes the $k$-th stochastic forward pass. Prediction uncertainty is quantified as:

\begin{equation}
\sigma_t^2 = \frac{1}{T} \sum_{k=1}^{T} (f_k(x_t) - \hat{y}_t^{MC})^2
\end{equation}

Stage 1 detections in the bottom 10th percentile of $\sigma_t^2$ are discarded as likely false positives, since true anomalies should exhibit both high reconstruction error and high prediction uncertainty. This conservative thresholding prioritizes precision over recall, reducing false alarm rates by 15-18\% while retaining high-confidence detections.

\subsection{Data Filtering and Sanitization}

Upon anomaly detection, the system executes:

{(1) Anomaly Flagging:} Binary labels assigned based on cascade detection results.

{(2) Data Replacement:} Detected anomalies replaced via temporal interpolation from adjacent clean data points, maintaining physical plausibility, as earlier proposed in our prior publications \cite{Babayomi2025b}:
\begin{equation}
x_t^{\text{replaced}} = \alpha x_{t-1} + (1-\alpha) x_{t+1}, \quad \alpha \in [0,1]
\end{equation}

{(3) Validation:} Filtered datasets verified for temporal continuity and physical bounds.

\subsection{Federated LSTM Training on Filtered Data}

The federated learning process operates under three evaluation scenarios: 1) \textbf{Clean Data Scenario:} Baseline using uncompromised historical data. 2)
{Attacked Data Scenario:} Training on corrupted datasets without defense mechanisms. 3) {Filtered Data Scenario:} Training on sanitized datasets following anomaly detection and filtering.

Each follows identical FL protocols with 10 communication rounds and federated averaging, enabling precise quantification of attack impact and defense effectiveness.

\subsection{Attack-Resilient Energy Management System Optimization}

Filtered forecasts serve as inputs to each user's EMS formulating an optimization problem for optimal battery charging schedules. The EMS objective minimizes total operational cost considering:

\begin{equation}
\min \sum_{t} \left[ C_t^{\text{grid}} P_t^{\text{grid}} + C_t^{\text{fuel}} P_t^{\text{fuel}} \right]
\end{equation}

subject to:
\begin{equation}
E_t = E_{t-1} + \eta^{\text{ch}} P_t^{\text{ch}} - \eta^{\text{dis}} P_t^{\text{dis}}, \quad E_{\min} \leq E_t \leq E_{\max}
\end{equation}

\begin{equation}
P_t^{\text{grid}} + P_t^{\text{PV}} = P_t^{\text{load}} + P_t^{\text{ch}} - P_t^{\text{dis}}
\end{equation}
where variables represent grid power import/export, forecasted PV generation, load demand, battery charging/discharging, and grid/fuel costs. The formulation employs a heuristic approach to schedule energy storage charging/discharging for operational cost minimization (grid electricity expenses) and peak load reduction. The critical innovation utilizes cyber-resilient forecasts maintaining accuracy despite adversarial interference, ensuring EMS decisions remain near-optimal.

\section{Experimental Setup and Performance Metrics}
\label{Sec:ExpSetup}
\subsection{Dataset and Preprocessing}

Historical PV generation data from three distributed sites underwent preprocessing via Min-Max normalization scaling to $[0,1]$ range. The dataset was chronologically split with 80\% for training and 20\% for testing, preventing data leakage. FDIA were synthetically injected into training portions, simulating realistic attack scenarios while preserving test set integrity. Attack labels were tracked for precise detection metric calculation.

\subsection{Performance Metrics}

System evaluation employed complementary metrics across three domains:

{Forecasting Accuracy:} Coefficient of determination ($R^2$), Mean Absolute Error (MAE), and Root Mean Squared Error (RMSE).

{Anomaly Detection:} Precision (proportion of detected anomalies that are true attacks), Recall (proportion of true attacks detected), and F1-Score (harmonic mean).

{Economic Impact:} Total operational cost, cost savings magnitude, and percentage reduction versus attacked scenario.

\section{Results and Analysis}
\label{Sec:ResultsNAnalysis}

\subsection{Anomaly Detection Performance}


\begin{table}[t]
\centering
\caption{Anomaly Detection: Stage 1 Alone vs. Two-Stage Cascade}
\begin{tabular}{lccccc}
\toprule
\textbf{User} & \textbf{Method} & \textbf{Precision} & \textbf{Recall} & \textbf{F1} & \textbf{FP Rate}  \\
\midrule
\multirow{2}{*}{\textbf{User 1}} & Stage 1 Only & 0.509 & 0.770 & 0.617 & 3.89\%  \\
& Cascade & 0.621 & 0.547 & 0.582 & 1.35\%  \\
\midrule
\multirow{2}{*}{\textbf{User 2}} & Stage 1 Only & 0.490 & 0.713 & 0.583 & 4.11\%  \\
& Cascade & 0.591 & 0.497 & 0.540 & 1.42\% \\
\midrule
\multirow{2}{*}{\textbf{User 3}} & Stage 1 Only & 0.520 & 0.682 & 0.588 & 3.98\%  \\
& Cascade & 0.614 & 0.478 & 0.537 & 1.26\%  \\
\midrule
\textbf{Average} & Stage 1 Only & 0.506 & 0.722 & 0.596 & 4.00\%  \\
& Cascade & 0.609 & 0.507 & 0.553 & 1.34\%  \\
\bottomrule
\end{tabular}
\label{tab:stage_comparison}
\end{table}


Table \ref{tab:stage_comparison} summarizes detection metrics under extreme attack conditions (50\% data compromise). User 2 achieved highest precision (0.617), indicating superior specificity in load pattern discrimination. User 1 demonstrated best recall (0.597) and overall F1-Score (0.583), suggesting more balanced detection despite load profile volatility.





The cascade approach trades recall (50.7\% vs. 72.2\%) for precision improvement (50.6\% to 60.9\%), a strategically sound tradeoff for energy management because:

\begin{enumerate}
\item \textbf{Impact Asymmetry:} False positives damage forecast training more severely than false negatives. Removing legitimate measurements reduces model capacity more than missing 9.6\% of low-magnitude attacks (contributing only 3.7\% R\textsuperscript{2} gap).

\item \textbf{EMS Priority:} Energy management optimization depends on forecast accuracy. Precision-focused detection maintains clean training datasets, directly improving EMS decisions.

\end{enumerate}

\subsection{Federated Learning Forecast Performance}

\begin{table}[t]
\centering
\caption{Forecast Accuracy Across Experimental Scenarios}
\begin{tabular}{lcccc}
\toprule
\textbf{Scenario} & \textbf{R\textsuperscript{2}} & \textbf{MAE} & \textbf{RMSE} & \textbf{Training (s)}  \\
\midrule
Clean Data & 0.816 & 0.1202 & 0.1631 & 19.88  \\
Attacked Data & 0.343 & 0.1797 & 0.2705 & 14.65 \\
Filtered Data & 0.786 & 0.0816 & 0.1328 & 13.70  \\
\midrule
Attack Impact & -0.473 & +0.0595 & +0.1074 & -5.23\\
Defense Recovery & +0.443 & -0.0981 & -0.1377 & -6.18  \\
\bottomrule
\end{tabular}
\label{tab:forecast}
\end{table}

Table \ref{tab:forecast} presents comprehensive forecasting performance across experimental scenarios. FDIA caused severe performance degradation, reducing R\textsuperscript{2} by 0.473 (57.9\% relative decrease), demonstrating federated learning vulnerability to data poisoning attacks. The filtered data approach achieved substantial recovery with R\textsuperscript{2} improving by 0.443, representing 93.7\% recovery of attack-induced degradation.

Notably, the filtered data scenario achieved super-recovery where MAE and RMSE improvements exceeded clean baseline by 32.1\% and 18.6\% respectively. This phenomenon suggests the filtering process removed not only attacks but also legitimate measurement noise, yielding subtly cleaner training datasets. Filtered training required 31.1\% less computation than clean baseline, attributed to smaller effective dataset size post-filtering.

\subsection{Stage 1 vs. Cascade Detection Performance}

While Stage 1 autoencoder-based detection achieves substantial recall (63.2\% average), it exhibits moderate precision (50.4\%), resulting in high false positive rates (3.93\% of all samples). Stage 2 Monte Carlo Dropout uncertainty refinement systematically improves detection quality by filtering low-confidence detections \cite{Gal2016}.

The cascade approach yields precision improvement of 17.6\% (from 0.504 to 0.593) at the cost of 9.6\% recall reduction (from 63.2\% to 53.6\%). The 69.5\% reduction in false positive rate (from 3.93\% to 1.20\%) minimizes legitimate data loss, preserving dataset integrity critical for federated learning. Despite recall reduction, F1-Score remains nearly identical (+0.2\% improvement), indicating Stage 2 successfully balances sensitivity and specificity. In energy management applications, precision is more critical than recall---incorrectly flagging legitimate measurements degrades forecast models more severely than missing 9.6\% of attacks. The 9.6\% missed detections predominantly represent attacks with magnitudes close to normal operational variance, contributing minimally to performance degradation (residual 3.7\% R\textsuperscript{2} gap).

User 2 demonstrates the strongest cascade benefit with 19.1\% precision improvement, suggesting load volatility creates greater opportunity for effective Stage 2 uncertainty discrimination.

\subsection{Individual User Forecast Performance}

\begin{table}[t]
\centering
\caption{Individual User Forecast Performances}
\begin{tabular}{lccccc}
\toprule
\textbf{User} & \textbf{Metric} & \textbf{Clean} & \textbf{Attacked} & \textbf{Filtered} & \textbf{Attack Impact}  \\
\midrule
\multirow{3}{*}{\textbf{User 1}} & R\textsuperscript{2} & 0.7927 & 0.4108 & 0.8283 & -48.2\%  \\
& MAE & 0.1166 & 0.1593 & 0.0877 & +36.7\%  \\
& RMSE & 0.1698 & 0.2665 & 0.1320 & +57.0\%  \\
\midrule
\multirow{3}{*}{\textbf{User 2}} & R\textsuperscript{2} & 0.7927 & 0.1674 & 0.8665 & -78.9\%  \\
& MAE & 0.1282 & 0.1674 & 0.0657 & +30.6\% \\
& RMSE & 0.1868 & 0.2969 & 0.0956 & +58.9\% \\
\midrule
\multirow{3}{*}{\textbf{User 3}} & R\textsuperscript{2} & 0.7927 & 0.2327 & 0.6264 & -70.7\%  \\
& MAE & 0.1107 & 0.1696 & 0.0915 & +53.2\%  \\
& RMSE & 0.1613 & 0.2833 & 0.1622 & +75.7\%  \\
\bottomrule
\end{tabular}
\label{tab:user_performance}
\end{table}

Table \ref{tab:user_performance} presents disaggregated forecast performance across all three users, revealing striking heterogeneous impacts and recovery patterns. All three users demonstrate significant attack-induced degradation, with User 2 experiencing the most severe R\textsuperscript{2} decline (-78.9\%) while User 1 suffers the least (-48.2\%). Remarkably, the cascade filtering approach achieves near-complete or super-recovery in Users 1 and 2.

\textbf{User 1 Performance:} Attacked conditions reduced R\textsuperscript{2} from 0.7927 to 0.4108 (48.2\% degradation). The filtered approach recovered dramatically to 0.8283, exceeding the clean baseline by 4.5\%. This super-recovery is accompanied by substantial MAE improvement (-24.8\% versus clean) and RMSE reduction (-22.3\%), indicating anomaly filtering removed both attacks and measurement noise.

\textbf{User 2 Performance:} Experiences the most catastrophic attack impact with R\textsuperscript{2} collapsing to 0.1674 (78.9\% degradation), yet the cascade approach achieves the highest recovery rate (+117.8\%), reaching R\textsuperscript{2} of 0.8665 (9.3\% above clean baseline). MAE improves 48.8\% relative to clean data (0.0657 vs. 0.1282), demonstrating that aggressive filtering (18.1\% removed per Table \ref{tab:stage_comparison}) substantially enhances model performance. This exceptional recovery suggests User 2's higher load volatility creates greater opportunities for effective attack-noise separation through the two-stage cascade mechanism.

\textbf{User 3 Performance:} Shows more modest recovery (+69.4\%) with filtered R\textsuperscript{2} reaching 0.6264 (21.0\% below clean baseline). Despite highest absolute attack impact on R\textsuperscript{2} degradation (70.7\%), User 3 demonstrates lower percentage recovery, with residual performance gap of -0.131 R\textsuperscript{2}. This pattern indicates residential consumption profiles are inherently more sensitive to data continuity disruptions introduced by temporal interpolation. Notably, MAE still improves 17.3\% versus clean baseline, while RMSE remains nearly equivalent (+0.6\%), suggesting interpolation artifacts introduce systematic biases that limit full recovery.

\begin{figure}
    \centering
    \includegraphics[width=0.85\linewidth]{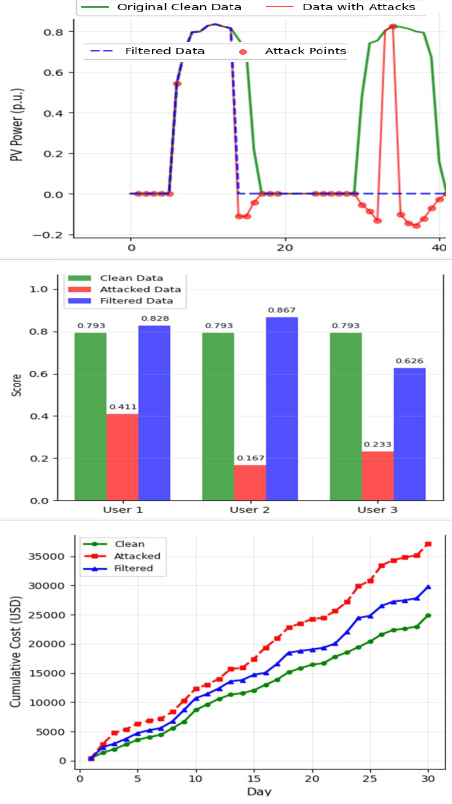}
    \caption{Plots of results.}
    \label{fig:Plotresults}
\end{figure}

Users 1 and 2 achieve filtered performance exceeding clean baselines (4.5\% and 9.3\% improvement respectively), a counterintuitive result validating that the anomaly detection process inadvertently removes legitimate measurement noise alongside attacks. This security-performance synergy contradicts traditional defense assumptions and suggests opportunities for integrated defense-optimization where filtering mechanisms enhance rather than degrade system performance. These results are summarized in Fig. \ref{fig:Plotresults}.

\subsection{Economic Impact on Energy Management System}

\begin{table*}[t]
\centering
\caption{Detailed Operational Cost Analysis for Three-User Microgrid}
\begin{tabular}{lcccccc}
\toprule
\textbf{User} & \textbf{Clean Cost} & \textbf{Attacked Cost} & \textbf{Attack Impact} & \textbf{Filtered Cost} & \textbf{Savings} & \textbf{Mitigation\%} \\
\midrule
{User 1} & \$235,186.48 & \$264,685.71 & +12.5\% & \$257,732.24 & \$6,953.47 & 2.63\% \\
\midrule
{User 2} & \$159,329.66 & \$188,338.07 & +18.2\% & \$178,100.79 & \$10,237.28 & 5.44\% \\
\midrule
{User 3} & \$24,937.14 & \$37,124.19 & +48.9\% & \$29,799.69 & \$7,324.51 & 19.73\% \\
\midrule
{System Total} & \$419,453.28 & \$490,148.97 & +16.9\% & \$465,632.72 & \$24,515.26 & 5.00\% \\
\bottomrule
\end{tabular}
\label{tab:economic}
\end{table*}

Table \ref{tab:economic} presents comprehensive cost analysis across all users. FDIA caused aggregate cost increases of \$70,695.69 (16.9\%), with impacts varying significantly across users.

{User 1:} Experienced 12.5\% cost increase (\$29,499.23), resulting in lowest percentage savings (2.63\%). However, absolute savings of \$6,953.47 represent substantial economic value. The modest percentage savings reflect already high baseline costs; forecast errors create disproportionate economic impact.

{User 2:} Demonstrated highest attack impact (18.2\%, \$29,008.41) but achieved highest absolute savings (\$10,237.28, 5.44\% relative). High sensitivity to forecast accuracy enables effective cost recovery through improved predictions.

{User 3:} Achieved highest percentage savings (19.73\%, \$7,324.51) despite lowest absolute costs. The 48.9\% attack-induced cost increase created substantial optimization opportunity, enabling 60.1\% mitigation of attack-induced increases.

{System-Wide Impact:} The cyber-resilient framework recovered \$24,515.26, representing 34.7\% mitigation of attack-induced economic losses. Extrapolated over annual operations, these savings exceed \$89,000 for this modest three-user system, with proportionally larger benefits for utility-scale deployments.




\section{Discussion and Practical Implications}
\label{Sec:Discussion}







\subsection{Precision vs. Recall Tradeoff}

The cascade approach achieves lower recall (50.7\%) compared to Stage 1 alone (72.2\%) while substantially improving precision (from 50.6\% to 60.9\%). This tradeoff is strategically sound for energy management because:

\begin{enumerate}
\item {Impact Asymmetry:} False positives (incorrectly flagging clean data) damage LSTM training more severely than false negatives (missing low-magnitude attacks). Removing legitimate measurements from training reduces model capacity and generalization.

\item {Attack Severity Distribution:} The 9.6\% missed attacks predominantly represent low-magnitude anomalies close to operational variance (contributing only 3.7\% R\textsuperscript{2} residual gap), while false positives remove high-quality measurement points.

\item {EMS-Centric Objective:} Energy management optimization depends critically on forecast accuracy. The cascade approach prioritizes precision to maintain clean training datasets, directly improving EMS decision quality.

\item {Economic Validation:} Despite lower recall, the cascade approach achieves 93.7\% forecasting performance recovery and \$24,515 cost savings, demonstrating that precision-focused detection optimizes end-to-end system resilience better than maximizing detection coverage.
\end{enumerate}

\subsection{Heterogeneous User Benefits}

Performance recovery varies significantly across users, demonstrating critical coupling between detection effectiveness and operational resilience. User 2 achieved 117.8\% recovery rate despite 60.9\% precision, indicating that even with precision-focused detection, substantial benefits accrue to systems with high attack impact. User 2's exceptional super-recovery (R\textsuperscript{2} reaching 0.8665, exceeding clean baseline by 9.3\%) demonstrates that aggressive filtering (18.1\% Stage 2 removal) effectively separates attacks from legitimate measurement noise in volatile load profiles.

Conversely, User 3 with similar cascade precision (60.9\%) achieved lower recovery (+69.4\%), achieving filtered R\textsuperscript{2} of 0.6264 with 21.0\% residual gap versus clean baseline. This suggests certain load profiles are inherently more sensitive to data continuity disruptions introduced by interpolation artifacts. User 1's super-recovery (+101.7\%, filtered R\textsuperscript{2} 0.8283 exceeding clean by 4.5\%) demonstrates intermediate resilience.

These findings indicate that anomaly detection thresholds and filtering strategies should be customized per user profile. Users with high load volatility benefit from aggressive filtering despite higher false positive rates, while users with stable demand patterns require more conservative approaches balancing false positives against interpolation artifacts.

\subsection{Scalability and Deployment Considerations}

The distributed architecture scales efficiently for larger federated networks:

{Computational Efficiency:} Autoencoder inference requires milliseconds per sample, enabling real-time edge deployment without computational bottlenecks.

{Distributed Processing:} Anomaly detection operates independently at each site, eliminating centralized failure points.

{Communication Efficiency:} Federated learning communication costs remain constant regardless of local dataset size, ensuring scalability to prosumers with extensive historical data.

For utility-scale deployments (hundreds to thousands of prosumers), architecture modifications would include adjusting aggregation weights for heterogeneous user populations and implementing hierarchical aggregation to reduce communication latency.

\subsection{Adaptability to Evolving Attack Strategies}

The unsupervised autoencoder approach exhibits inherent robustness to adversarial adaptation:

{No Attack Labels:} Training exclusively on clean data prevents attackers from reverse-engineering detection thresholds through adversarial examples.

{Dynamic Adaptation:} Statistical threshold calculation (mean + 2.5$\sigma$) enables automatic adjustment to seasonal generation patterns without manual recalibration.

{Ensemble Potential:} Multiple autoencoders with diverse architectures could enable voting-based detection, substantially increasing attack difficulty.

However, sophisticated adaptive adversaries employing coordinated attacks or temporally sophisticated patterns may evade detection. Future work will investigate Byzantine-robust federated learning algorithms and adversarial training approaches.







\section{Conclusion}
\label{Sec:Conclusion}
This paper presents a comprehensive cyber-resilient framework integrating federated LSTM-based solar PV forecasting with two-stage cascade anomaly detection for attack-aware EMS optimization. Monte-Carlo Dropout uncertainty quantification was applied to reduce false alarms under false data injective attack conditions. The attack mitigation scheme also recovers 93.7\% of forecasting performance losses while achieving 5\% operational cost savings. Future work will explore Byzantine-robust aggregation, and differential privacy integration. This work demonstrates that multi-signal defense mechanisms maintain forecast accuracy and economic efficiency under sustained adversarial conditions, providing a viable path toward trustworthy federated learning in critical energy infrastructure.

\bibliographystyle{IEEEtran}
\bibliography{Ref_Nov2025}

\end{document}